УДК 004.75

**М.І. Козленко**, канд. техн. наук, доц.

Державний вищий навчальний заклад „Прикарпатський національний університет імені Василя Стефаника", м.Івано-Франківськ, Україна, e-mail: kozlenkomykola@ukr.net

# ЗАВАДОСТІЙКІСТЬ ПЕРЕДАЧІ ДАНИХ У СИСТЕМАХ ОПРАЦЮВАННЯ ТЕЛЕМЕТРИЧНОЇ ІНФОРМАЦІЇ АВТОНОМНИХ МОБІЛЬНИХ РОБОТІВ

**M.I. Kozlenko**, Cand. Sci. (Tech.), Assoc. Prof.

State Higher Educational Institution "Vasyl Stefanyk Precarpathian National University", Ivano-Frankivsk, Ukraine, e-mail: kozlenkomykola@ukr.net

# THE INTERFERENCE IMMUNITY OF THE TELEMETRIC INFORMATION DATA EXCHANGE WITH AUTONOMOUS MOBILE ROBOTS

**Мета**. Отримати залежність імовірності спотворення двійкового символу від нормованого відношення сигнал/шум при обміні даними сигналами зі змінною ентропією в розподілених системах оброблення телеметричної інформації.

**Методика**. Результати отримані шляхом теоретичної розробки та підтверджені моделюванням в обчислювальному експерименті.

**Результати**. Отримана залежність імовірності спотворення двійкового символу від нормованого відношення сигнал/шум для методу обміну даними за допомогою широкосмугових сигналів зі змінною ентропією. Встановлено, що завадостійкість методу є меншою на 4,5дБ у порівнянні з оптимальним опрацюванням ортогональних сигналів. Проте, часова складність є суттєво меншою, що дозволяє опрацьовувати сигнали з більшою базою та в умовах однакової часової складності покращити завадостійкість на 2дБ при реалізації пристрою оброблення на базі 8-ми розрядного мікроконтролера.

**Наукова новизна**. Уперше отримана аналітична залежність імовірності спотворення двійкового символу від нормованого відношення сигнал/шум при обміні даними сигналами зі змінною ентропією.

**Практична значимість**. Отримані результати дозволяють обґрунтувати доцільність застосування методу обміну даними сигналами зі змінною ентропією в розподілених системах оброблення телеметричної інформації в конкретних умовах.

**Ключові слова**: *широкосмуговий сигнал, змінна ентропія, завадостійкість, телеметрія, мобільний робот*

**Вступ.** Розвиток сучасних інформаційних технологій усе частіше вимагає застосування територіально розподіленого опрацювання інформації. Існує велика кількість застосувань, що вимагають надійної передачі даних, живучості систем у складних умовах за дії інтенсивних завад природного, техногенного та іншого походження за низьких відношень сигнал/шум з відстанями порядку одиниць або десятків кілометрів. Зокрема, це системи передачі та опрацювання телеметричної інформації від автономних мобільних роботів, де вимоги щодо швидкості обміну даними, як правило, можуть бути знижені до величин порядку $10^2$ до $10^3$ біт/с.

Умовою створення якісних розподілених систем оброблення інформації є побудова простих та надійних засобів обміну даними, у тому числі й бездротових. Найчастіше задачі створення згаданих систем вирішуються шляхом побудови традиційних каналів обміну даними, в яких застосовуються прості сигнали-носії з малою базою, найпоширенішими з яких є сигнали безпосереднього низькочастотного представлення дискретних повідомлень та гармонійні сигнали, модульовані чи маніпульовані у відповідності до вихідного повідомлення. Проте не завжди такий підхід задовольняє сучасним вимогам, зокрема, за такими показниками як надійність, простота апаратної реалізації, стабільність характеристик, можливість обміну даними при малих відношеннях сигнал/шум тощо. Одним з перспективних шляхів розвитку обміну даними є використання сигналів з великою базою, зокрема, з розширеним спектром, що може стати основою забезпечення надійності функціонування каналоутворюючого обладнання та високої стабільності характеристик інформаційних систем в експлуатаційних умовах.

**Постановка проблеми.** Необхідність організації надійного обміну даними в розподілених системах оброблення інформації зумовлює практичне завдання зі створення простих, надійних та недорогих каналоутворюючих пристроїв. Результативне вирішення цього завдання можливе за умови успішного розв'язання наукових проблем створення та розвитку нових ефективних методів передавання та приймання інформації, зокрема, способів формування та обробки широкосмугових сигналів.

**Аналіз останніх досліджень та публікацій.** На даний час розроблено багато способів передавання та приймання інформації з використанням широкосмугових сигналів. Більшість із них базується на використанні додаткового сигналу, що розширює спектр, при передаванні та на подальшому кореляційному опрацюванню на стороні приймання. Суть такого опрацювання полягає в порівнянні форми розширюючого спектр сигналу зі зразком, що зберігається в пам'яті приймального пристрою. Такий підхід дозволяє забезпечити ефективність роботи системи обміну даними та отримати інші переваги, що є характерними для систем, які реалізуються на основі такого підходу [1]. Проте реалізація згаданих систем є складною технічною задачею, а приймально-передавальні пристрої, як правило, потребують значного апаратного та програмного ресурсу. Доцільно зазначити, що сигнали, які використовуються для розширення спектру в таких випадках, наприклад, псевдовипадкові послідовності, мають характеристики, які лише наближаються до випадкових, що не дозволяє використати всі можливості випадкових сигналів і призводить до необхідності зберігання відповідних взірців форми сигналів на стороні приймання. Крім того, необхідність зменшення ймовірності помилок у процесі обміну даними зумовлює зростання бази широкосмугових сигналів, і, як наслідок, призводить до зростання апаратних затрат, що не завжди є виправданим [1].

Існують також інші варіанти реалізації згаданих систем. Зокрема, такі, в яких взірець псевдовипадкового сигналу, що розширює спектр, не зберігається постійно на стороні приймання, а передається до приймача окремим паралельним каналом до початку основного сеансу обміну [1]. Значний розвиток мають способи обміну даними, в основі розширення спектру яких лежить використання негармонійних форм сигналів-носіїв [2]. Також відомі способи, де широкосмугові сигнали-носії формуються та обробляються за допомогою явищ динамічного детермінованого хаосу [3]. Знайшли поширення методи формування сигналів з частотним розділенням у бездротових системах [4].

Автору належить започаткування розв'язання проблеми шляхом використання широкосмугових сигналів зі змінною ентропією. Запропонований метод формування та опрацювання широкосмугових сигналів базується на використанні в якості носія шумоподібного випадкового сигналу, ентропія розподілу ймовірностей амплітудних значень якого поставлена у відповідність до символів інформаційного повідомлення, що передається. На даний час проведене дослідження впливу відновлення проміжних значень прийнятого сигналу на завадостійкість. Оцінена рівномірність розподілу енергії таких сигналів за частотами [5]. Проаналізована ефективність використання сигналів різних часових форм.

**Виділення невирішених раніше частин загальної проблеми.** Раніше невирішеною частиною загальної проблеми є отримання кількісних показників завадостійкості такого методу, саме цьому й присвячена дана робота.

**Формулювання цілей даної роботи.** Одним з невирішених раніше питань щодо способу передавання та приймання інформації за допомогою сигналів зі змінною ентропією є дослідження показників його завадостійкості. Отже, об'єктом дослідження є завадостійкість способу, а отримання її кількісних показників основною метою роботи.

**Метод обміну даними сигналами зі змінною ентропією.** Суть методу полягає у формуванні випадкового широкосмугового сигналу-носія таким чином, що його ентропія розподілу ймовірностей амплітудних значень або станів, якщо сигнал дискретний, поставлена у відповідність до символів інформаційного повідомлення. Для випадку двійкового базису сигналу повідомлення, це відбувається у спосіб, коли один з дискретних символів, наприклад, „1", представлений випадковим сигналом $s_1(t)$ з певним значенням ентропії, а другий, відповідно, „0", випадковим сигналом $s_2(t)$ з іншим значенням ентропії.

Значення ентропії довільного сигналу $x(t)$ визначається наступним чином [1]

$$H_{x(t)} = -\sum_{j=1}^{m} p(X_j) \cdot \log_2(p(X_j)), \quad (1)$$

де $j$ – порядковий номер стану (значення амплітуди) сигналу; $m$ – загальна кількість дискретних станів сигналу; $X_j$ – значення стану з порядковим номером $j$; $p(X_j)$ – імовірність стану $X_j$. При цьому вважається, що $0 \cdot \log_2(0) = 0$.

Сигнал у каналі є неперервним, але формується та обробляється у цифровому представленні з використанням відповідних перетворень. Під станами сигналу, у даному випадку, слід розуміти його квантовані амплітудні значення. У межах проведеного дослідження використовувалися 16-ти бітові аналого-цифрові (АЦП) та цифро-аналогові перетворювачі (ЦАП), отже сигнал розглядається як такий, що має $m = 65536$ станів. Ентропія таких сигналів може приймати значення від 0 до 16біт/відлік.

Проведені раніше дослідження показують ефективність застосування сигналів-носіїв для цього методу у вигляді випадкового сигналу з розподілом імовірностей станів близьким до нормального та рівномірною спектральною щільністю енергії в межах робочої смуги частот.

При прийманні відбувається опрацювання прийнятого сигналу $r(t)$. Сигнал $r(t)$ є сумою згортки переданого сигналу ($s_1(t)$ або $s_2(t)$) з імпульсною характеристикою каналу $h_c(t)$ та завади $n_{кан}(t)$, джерелом якої є канал зв'язку. Обробка полягає у статистичному оцінюванні ентропії його послідовних фрагментів $^k r(t)$, що відповідають $k$-тому символьному інтервалу повідомлення, які можна розглядати як $k$-ту реалізацію випадкового процесу, яким є прийнята суміш сигналу та завади. Оцінювання відбувається згідно з (2), оскільки амплітуди сигналу нормально розподілені, за кі-нцевим проміжком часу (часу символьного інтервалу) на підставі однієї реалізації, що є обґрунтованим для випадку стаціонарних та ергодичних процесів.

$$\hat{H}_{r(t)} = \log_2 \sqrt{2\pi e s^2_{r(t)}}, \qquad (2)$$

де $s^2_{r(t)}$ – статистична оцінка дисперсії прийнятого сигналу $r(t)$; $e \approx 2{,}718$ – основа натурального логарифма.

Згідно з [1], демодуляцією вважається виділення низькочастотного представлення повідомлення з сигналу-носія, а детектуванням – процес ухвалення рішення відносно значення прийнятого інформаційного символу. При прийманні сигналу, у моменти часу $t=T$, тобто в моменти завершення символьного інтервалу, на виході демодулятора, у переддетекторній точці, формується сигнал $z(T)$, значення якого є випадковою величиною та містять у собі випадкову флуктуацію $n_0(T)$ ($n_{01}(T)$ та $n_{02}(T)$ при опрацюванні „1" і „0" відповідно) з дисперсією $\sigma_0^2$ ($\sigma_{01}^2$ та $\sigma_{02}^2$ при опрацюванні „1" і „0" відповідно), яка обумовлена відхиленнями від стаціонарності завад у каналі, похибкою статистичного оцінювання ентропії, обмеженою в часі тривалістю сигналів, випадковим характером самих сигналів і, відповідно, вибірок з нього, фактично, є завадою, що діє на виході демодулятора. Раніше проведеними дослідженнями встановлено нормальний характер розподілу цієї завади, а також симетрію функцій правдоподібності, тобто $\sigma_0^2 = \sigma_{01}^2 = \sigma_{02}^2$. Для даного методу значення сигналу $z(T)$ визначається як статистична оцінка ентропії $\hat{H}_{r(t)}$ суміші корисного сигналу та завади на вході демодулятора за кінцевим проміжком часу. У випадку, коли передається сигнал $s_1(t)$, математичне сподівання $z(T)$ дорівнює $a_1$, а дисперсія $\sigma_{01}^2$, в іншому випадку, коли передається сигнал $s_2(t)$, математичне сподівання $z(T)$ дорівнює $a_2$, а дисперсія $\sigma_{02}^2$ відповідно. Таким чином, $a_1$ та $a_2$ – бажані сигнальні компоненти на виході демодулятора, а дисперсії $\sigma_{01}^2$ та $\sigma_{02}^2$ – характеризують потужність шумової компоненти-завади $n_{01}(T)$ або $n_{02}(T)$ на виході демодулятора [1].

Процес детектування інформаційних символів полягає у визначенні приналежності прийнятого сигналу до однієї з двох (для випадку двійкового базису повідомлень) областей. Це відбувається шляхом порівняння сигналу $z(T)$ у момент завершення символьного інтервалу з порогом $\gamma_0$, що розраховується, виходячи з критерію мінімізації ймовірності прийняття хибного рішення щодо значення прийнятого символу, згідно з (3). Таке значення порогу є оптимальним за умови рівної ймовірності появи інформаційних символів та симетрії функцій правдоподібності сигналів [1].

$$\gamma_0 = \frac{a_1 + a_2}{2}. \qquad (3)$$

**Викладення основного матеріалу досліджень.** Для даного дослідження вибір сигналів $s_1(t)$, $s_2(t)$ та завади $n_{кан}(t)$ здійснено в один з найбільш наочних варіантів, коли перший стан інформаційного символу, „1" – випадковий сигнал $s_1(t)$ з розподілом імовірностей близький до нормального, рівномірною спектральною щільністю та відповідним сталим рівнем ентропії, а другий стан символу, „0" – сигнал $s_2(t)$ – пасивна пауза, за якої відсутня активність передавача та потужність не випромінюється, з нульовим рівнем ентропії. Завада $n_{кан}(t)$, що діє в каналі, розглядається як стаціонарний адитивний білий гаусів шум – AWGN.

Як відомо з [6], мірою завадостійкості обміну дискретними повідомленнями є залежність імовірності спотворення символів від відношення енергії сигналу до спектральної щільності потужності завади, а також залежність відношення сигнал/шум у точці ухвалення рішень від такого відношення на вході пристрою оброблення. У випадку використання двійкового базису повідомлень, перша з цих залежностей – залежність імовірності спотворення двійкового символу $P_b$ від нормованого відношення сигнал/шум $E_b/N_0$, де $E_b$ – середня енергія, що припадає на один оброблюваний біт, на вході пристрою оброблення; $N_0/2$ – двобічна спектральна щільність потужності завади на вході цього пристрою.

Імовірність спотворення двійкового символу, для випадку нормально розподіленої завади в точці ухвалення рішень, визначається [1] згідно з виразом

$$P_b = Q\left(\frac{a_1 - a_2}{2\sigma_0}\right), \qquad (4)$$

де $a_1$ – математичне сподівання сигналу $z(T)$ у випадку оброблення сигналу $s_1(t)$ (перша сигнальна компонента); $a_2$ – математичне сподівання сигналу $z(T)$ у випадку оброблення сигналу $s_2(t)$ (друга сигнальна компонента); $\sigma_0^2$ – дисперсія нормально розподіленої завади $n_0(T)$ на виході демодулятора, у точці ухвалення рішень; $Q(x)$ – гаусів інтеграл помилок, що визначається згідно з виразом (5)

$$Q(x) = \frac{1}{\sqrt{2\pi}} \int_x^\infty \exp\left(-\frac{u^2}{2}\right) du, \qquad (5)$$

де $u$ – допоміжна змінна.

Статистична оцінка ймовірності спотворення двійкового символу визначається згідно з наступним виразом, шляхом підстановки у (4) значень відповідних статистичних оцінок замість істинних значень $a_1$, $a_2$ та $\sigma_0$.

$$\hat{P}_b = Q\left(\frac{\hat{a}_1 - \hat{a}_2}{2s_0}\right), \qquad (6)$$

де $\hat{a}_1$ – статистична оцінка величини $a_1$; $\hat{a}_2$ – статистична оцінка величини $a_2$; $s_0^2$ – оцінка дисперсії $\sigma_0^2$ AWGN завади $n_0(T)$ у точці ухвалення рішень.

Визначення ймовірності помилок та її оцінки згідно з (4), (6) буде коректним за умови симетрії функцій правдоподібності, що має місце в даному випадку. У свою чергу, відношення $E_b/N_0$ визначається згідно з наступним виразом

$$\frac{E_b}{N_0} = \frac{S}{N} \cdot \frac{W}{R}, \qquad (7)$$

де $S$ – середня потужність сигналу на вході пристрою оброблення; $N$ – середня потужність завади на вході пристрою оброблення; $W$ – ширина частотного спектру; $R$ – швидкість обміну даними.

Середня потужність $S$ сигналу при однаковій частоті появи та тривалості $s_1(t)$ та $s_2(t)$ визначається згідно з наступним виразом (для розглянутих сигналів потужність на вході $S_2 = 0$ Вт, оскільки передавач неактивний під час пасивної паузи)

$$S = \frac{S_1 + S_2}{2}, \qquad (8)$$

де $S_1$ – потужність сигналу $s_1(t)$; $S_2$ – потужність сигналу $s_2(t)$.

Аналітичний вираз для залежності ймовірності спотворення двійкового символу від $E_b/N_0$ можна отримати, виходячи з наступного.

Сигнальна компонента $a_1$ при обробленні сигналу $s_1(t)$ визначається за таким виразом

$$a_1 = \log_2 \sqrt{2\pi e(S_1 + N)}. \qquad (9)$$

Сигнальна компонента $a_2$ при обробленні сигналу $s_2(t)$ визначається за таким виразом

$$a_2 = \log_2 \sqrt{2\pi e N}, \qquad (10)$$

з урахуванням того, що $S_1 = 2S$, оскільки $S_2 = 0$, шляхом підстановки в (4) значень (9), (10) та (2), отримуємо

$$P_b = Q\left(\frac{\log_2\left(1 + 2\frac{S}{N}\right)}{4\sigma_0}\right). \qquad (11)$$

У свою чергу, для визначення $\sigma_0$ необхідно визначити дисперсію $\text{var}\{s^2_{r(t)}\}$ статистичної оцінки $s^2_{r(t)}$ дисперсії $\sigma^2_{r(t)}$ вхідного сигналу $r(t)$. Згідно з відомою формулою

$$\text{var}\{s^2_{r(t)}\} = \frac{1}{n}\left(\mu_4 - \frac{n-3}{n-1}\sigma^4_{r(t)}\right), \qquad (12)$$

де $n$ – розмір вибірки; $\mu_4$ – центральний момент 4-го порядку.

Центральний момент нормально розподіленої величини визначається за таким виразом

$$\mu_w = \begin{cases} 0, & w - \text{не парне} \\ \frac{w!}{(w/2)!} \cdot \left(\frac{\sigma^2_{r(t)}}{2}\right)^{w/2}, & w - \text{парне} \end{cases}. \qquad (13)$$

Отже, центральний момент 4-го порядку

$$\mu_4 = 3\sigma^4_{r(t)}. \qquad (14)$$

Підставляючи (14) у (12), отримуємо

$$\text{var}\{s^2_{r(t)}\} = \frac{2\sigma^4_{r(t)}}{n-1}. \qquad (15)$$

Слід зауважити, що оцінка дисперсії $s^2_{r(t)}$ є випадковою величиною, що розподілена за $\chi^2$ законом, однак при збільшенні $n$ цей розподіл асимптотично наближається до нормального і при $n \geq 30$ його можна вважати нормальним.

Оскільки оцінка $\hat{H}_{r(t)}$ ентропії вхідного сигналу пов'язана з оцінкою $s^2_{r(t)}$ його дисперсії детермінованою функцією (2), то для отримання дисперсії оцінки $\hat{H}_{r(t)}$ слід скористатися відомими формулами, що пов'язують щільність імовірностей випадкової величини та функцію перетворення з числовими характеристиками (математичне сподівання та дисперсія) її перетворення.

Як відомо, математичне сподівання оцінки $s^2_{r(t)}$ є $\sigma^2_{r(t)}$, дисперсія цієї оцінки визначається з (15). Отже, вважаючи розподіл нормальним, щільність імовірностей оцінки $s^2_{r(t)}$

$$f(s^2_{r(t)}) = \frac{1}{\sqrt{\frac{4\pi\sigma^4_{r(t)}}{n-1}}} \cdot \exp\left(-\frac{(s^2_{r(t)} - \sigma^2_{r(t)})^2}{\frac{4\sigma^4_{r(t)}}{n-1}}\right). \qquad (16)$$

Математичне сподівання оцінки $\hat{H}_{r(t)}$ визначається наступним чином

$$\text{E}\{\hat{H}_{r(t)}\} = \int_0^\infty \log_2 \sqrt{2\pi e s^2_{r(t)}} \cdot f(s^2_{r(t)}) ds^2_{r(t)}, \qquad (17)$$

а її дисперсія, відповідно,

$$\text{var}\{\hat{H}_{r(t)}\} = \int_0^\infty \left(\log_2 \sqrt{2\pi e s^2_{r(t)}}\right)^2 \cdot f(s^2_{r(t)}) ds^2_{r(t)} - \left(\text{E}\{\hat{H}_{r(t)}\}\right)^2, \qquad (18)$$

а, отже, шукане СКВ

$$\sigma_0 = \sqrt{\operatorname{var}\{\hat{H}_{r(t)}\}}. \qquad (19)$$

Очевидно, що обчислення $\sigma_0$ за (16)–(19) є суттєво ускладненим, оскільки містить інтегрування функції щільності ймовірностей нормального розподілу, і практично можливим із застосуванням числових методів. Тому слід розглянути можливість отримання спрощеного наближеного виразу для $\sigma_0$. Для спрощення проведемо заміну логарифмічної функції (2) на її лінійну апроксимацію таким чином, щоб у точці, що відповідає математичному сподіванню оцінки $s^2_{r(t)}$, значення апроксимації та її першої похідної дорівнювали значенню логарифмічної функції (2) та її першій похідній відповідно. Для нормального закону розподілу така заміна не приводить до суттєвого спотворення дисперсії, оскільки, з високою ймовірністю, значення випадкової величини зосереджені у вузькому інтервалі (правило трьох сигм) у порівнянні з її областю значень. Перша похідна функції (2) визначається за наступним виразом

$$\frac{d\hat{H}_{r(t)}}{ds^2_{r(t)}} = \left(\log_2 \sqrt{2\pi e s^2_{r(t)}}\right)' = \frac{0,5}{s^2_{r(t)} \cdot \ln 2} = \frac{0,721}{s^2_{r(t)}}. \quad (20)$$

У точці, що відповідає математичному сподіванню $\mathrm{E}\{s^2_{r(t)}\} = \sigma^2_{r(t)}$ оцінки $s^2_{r(t)}$, значення похідної складає $0,721/\sigma^2_{r(t)}$. Отже, враховуючи лінійність перетворення, отримуємо вираз для $\sigma_0$

$$\sigma_0 \approx \sqrt{\frac{2\sigma^4_{r(t)}}{(n-1)}} \cdot \frac{0,721}{\sigma^2_{r(t)}} \approx \frac{1,02}{\sqrt{n-1}}. \qquad (21)$$

Оцінено відхилення значення $\sigma_0$, отриманого теоретично за допомогою виразів (16)–(19) і визначених за спрощеною формулою (21). Встановлено, що для розміру вибірки у 1024 відліки відхилення не перевищує 0,26%.

**Основні результати досліджень.** Таким чином, підставляючи (21) в (11), наближена формула для ймовірності спотворення двійкового символу має наступний вигляд

$$P_b \approx Q\left(0,245 \cdot \log_2\left(1 + 2\frac{S}{N}\right) \cdot \sqrt{n-1}\right). \quad (22)$$

Для даного методу, кількість відліків протягом символьного інтервалу пов'язана з тривалістю символьного інтервалу та швидкістю обміну даними при заданій частоті дискретизації. Отже, при заданій ширині частотного спектру $W$, кількість відліків прямо пов'язана з базою сигналу. Оскільки $T = n/f_s$ і $f_s = 2W$ (дискретизація за Найквістом), то

$$R = \frac{1}{T} = \frac{2W}{n}; \qquad (23)$$

$$B = \frac{W}{R} = \frac{n}{2}. \qquad (24)$$

Підставляючи (24) в (7), отримуємо наступні вирази для зв'язку $E_b/N_0$ та $S/N$

$$\frac{E_b}{N_0} = \frac{S}{N} \cdot \frac{n}{2}; \qquad (25)$$

$$\frac{S}{N} = \frac{2E_b}{N_0 n}. \qquad (26)$$

Підставляючи (26) у (22), отримаємо залежність $P_b$ від $E_b/N_0$

$$P_b \approx Q\left(0,245 \cdot \log_2\left(1 + \frac{4E_b}{N_0 n}\right) \cdot \sqrt{n-1}\right). \quad (27)$$

З виразу (27) встановлено, що для даного методу значення кількості відліків у вибірці (а отже й бази сигналів) істотно впливає на ймовірність помилки при заданому відношенні $E_b/N_0$, що пояснюється залежністю потужності завади в точці ухвалення рішень від кількості відліків у вибірці. Крім того, встановлено, що залежність аргументу гаусового інтегралу помилок від $n$ не є монотонною. Для встановлення оптимального розміру вибірки знайдено значення $n$, що максимізує аргумент гаусового інтегралу помилок у (27) при постійному $E_b/N_0$ (у даному випадку обрано величину 20дБ з практичних міркувань). Встановлено, що в таких умовах цей максимум (а отже й мінімум імовірності помилки) досягається при $n=105$. Також проведене моделювання в обчислювальному експерименті, у ході якого побудована залежність оцінки обраного критерію завадостійкості $\hat{K} = (\hat{a}_1 - \hat{a}_2)/(2s_0)$ від $S/N$ та розміру вибірки $n$ (рис. 1). Використання в якості критерію завадостійкості $K$ (та його статистичної оцінки $\hat{K}$) аргументу гаусового інтегралу помилок можливе завдяки його однозначному зв'язку з імовірністю помилок.

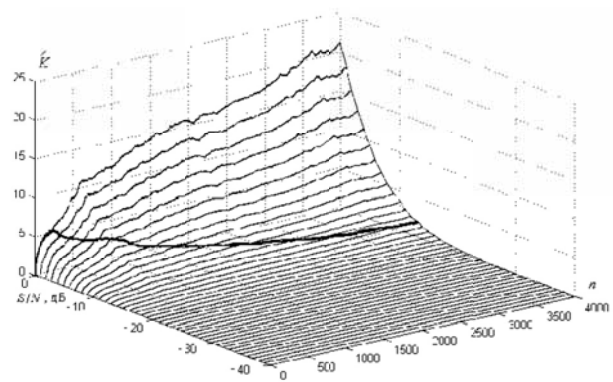

*Рис. 1. Залежність $\hat{K} = (\hat{a}_1 - \hat{a}_2)/(2s_0)$ від $n$ та $S/N$*

На думку автора, використання цієї величини в деяких випадках є більш наглядним. На поверхні виділе-но криву, що відповідає постійному $E_b/N_0$ (у даному випадку 20дБ). Ця крива представлена на рис. 2 в залежності тільки від однієї координати – розміру вибірки. Встановлено, що максимуму оцінка значення критерію досягає при розмірі вибірки у 10³ відліків. Отже, відхилення результатів, отриманих на основі встановлених аналітичних залежностей і отриманих при моделюванні, не перевищує 2%. Також на рис. 2 подані значення $K$, обчислені за (27) ($K$ – аргумент гаусового інтегралу помилок у (27)).

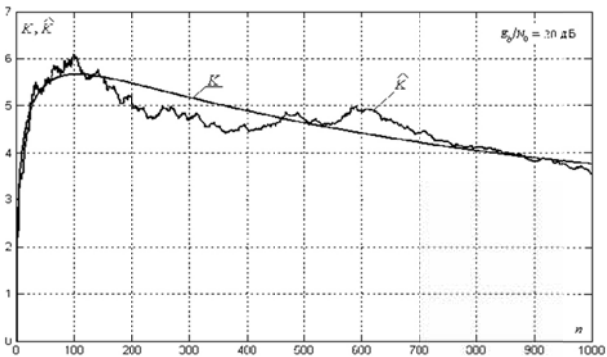

*Рис. 2. Залежність* $\hat{K} = \left(\hat{a}_1 - \hat{a}_2\right) / (2s_0)$ *від n при* $E_b/N_0 = 20дБ$

Отже, у подальшому розглядається розмір вибірки у 10⁵ відліків, що при обраній частоті дискретизації АЦП (48кСемпл/с), який використовувався для експериментальних досліджень, відповідає базі сигналу близько 17дБ, тривалості символьного інтервалу 2,1875мс та швидкості обміну даними близько 460біт/с. На основі отриманих залежностей побудовано криву завадостійкості для розробленого методу. На рис. 3 її подано для бази сигналу 17дБ (крива 2).

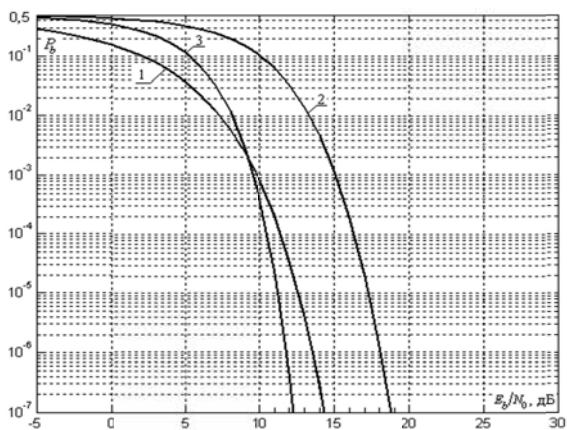

*Рис. 3. Залежність імовірності спотворення двійкового символу* $P_b$ *від відношення* $E_b/N_0$: *1 – оптимальне оброблення ортогональних сигналів; 2 – завадостійкість методу передачі сигналів з керованою ентропією* $(W/R = 17 \text{ dB})$; *3 – завадостійкість в умовах однакової часової складності алгоритмів опрацювання*

Побудову залежності $P_b$ від $E_b/N_0$ проведено шляхом обчислення $P_b$ згідно з (27) зі змінним відношенням $E_b/N_0$ при фіксованому значенні бази сигналу $W/R$. Також проведено моделювання в обчислювальному експерименті, за якого отримані оцінки згідно з (6), результати якого практично збіглись із теоретичними. Для порівняння, на рис. 3 також наведено характеристику завадостійкості для оптимального опрацювання ортогональних сигналів (крива 1), наприклад, FSK-FHSS. Імовірність спотворення двійкового символу для цього випадку визначається згідно з виразом [1]

$$P_b = Q\left(\sqrt{\frac{E_b}{N_0}}\right). \qquad (28)$$

Як можна побачити, розроблений метод за завадостійкістю наближається до оптимального кореляційного оброблення ортогональних сигналів з різницею, що не перевищує 4,5дБ, при ймовірності помилок на рівні 10⁻⁶. Раніше проведеними експериментальними дослідженнями встановлено [7], що програмне забезпечення низькорівневого оброблення сигналів, яке реалізоване на основі методу змінної ентропії, дозволяє опрацювати у 8,1 разів більше відліків сигналу, ніж при кореляційному обробленні на 8-ми розрядному мікроконтролері типу AVR (ATmega1284). Отже, вивільняється додатковий обчислювальний ресурс, що можна використати для покращення завадостійкості шляхом використання сигналів з більшою базою. Як можна побачити, для такого випадку, максимальне покращення завадостійкості складає величину близько 5дБ, а при зафіксованій на рівні 10⁻⁶ імовірності помилок – не менше 2дБ в умовах однакової часової складності алгоритмів опрацювання (крива 3, рис. 3 для 8-ми розрядного мікроконтролера). Також вивільнений обчислювальний ресурс можна спрямувати на реалізацію завадостійкого кодування, зокрема, турбо-кодами, що призводить до суттєвого покращення завадостійкості інформаційного обміну При опрацюванні сигналів за допомогою 16 та 32 розрядних мікроконтролерів перевага є значно меншою. Зокрема, збільшення кількості відліків, що опрацьовуються за однаковий проміжок часу для 16-ти розрядного мікроконтролера MSP430G2231, складає 2,26 рази, а для 32-розрядного STM32F100RBT6B – 2,72 рази.

**Висновки.** Завадостійкість методу обміну даними широкосмуговими сигналами з керованою ентропією є нижчою від теоретичної завадостійкості оптимального кореляційного оброблення ортогональних сигналів, проте реалізація апаратної частини приймально-передавального обладнання є значно простішою, а часова складність демодуляції є суттєво меншою, що дозволяє за однаковий проміжок часу обробляти сигнали з більшою базою, тим самим суттєво покращити завадостійкість. Однією з позитивних властивостей способу є також збереження працездатності за низьких відношень сигнал/шум.

**Перспективи подальших досліджень.** Основними напрямами подальшого дослідження є вдосконалення процедури оцінювання ентропії, з метою мінімізації помилки, пошук типів сигналів, що забезпечують вищу ефективність демодуляції, розробка ефективних способів демодуляції, реалізація способів ефективної бітової синхронізації тощо [8].

**Цель**. Получить зависимость вероятности искажения двоичного символа от нормированного отношения сигнал/помеха при обмене данными сигналами с переменной энтропией в распределенных системах обработки телеметрической информации автономных мобильных роботов.

**Методика**. Результаты получены путем теоретической разработки и подтверждены моделированием в вычислительном эксперименте.

**Результаты**. Получена зависимость вероятности искажения двоичного символа от нормированного отношения сигнал/помеха для метода обмена данными с помощью широкополосных сигналов с переменной энтропией. Установлено, что помехоустойчивость метода меньше на 4,5дБ в сравнении с оптимальной обработкой ортогональных сигналов. Однако, временная сложность существенно меньшая, что позволяет обрабатывать сигналы с большей базой и в условиях одинаковой временной сложности улучшить помехоустойчивость на 2дБ при реализации устройства обработки сигналов на базе 8-ми разрядного микроконтроллера.

**Научная новизна**. Впервые получена аналитическая зависимость вероятности искажения двоичного символа от нормированного отношения сигнал/шум при обмене данными сигналами с переменной энтропией.

**Практическая значимость**. Полученные результаты позволяют обосновать целесообразность применения метода обмена данными сигналами с переменной энтропией в распределенных системах обработки телеметрической информации в конкретных условиях.

**Ключевые слова:** *широкополосный сигнал, переменная энтропия, помехоустойчивость, телеметрия, мобильный робот*

**Purpose**. To obtain the interference immunity of the data exchange by spread spectrum signals with variable entropy of the telemetric information data exchange with autonomous mobile robots.

**Methodology.** The results have been obtained by the theoretical investigations and have been confirmed by the modeling experiments.

**Findings**. The interference immunity in form of dependence of bit error probability on normalized signal/noise ratio of the data exchange by spread spectrum signals with variable entropy has been obtained. It has been proved that the interference immunity factor (needed normalized signal/noise ratio) is at least 2 dB better under condition of equal time complexity as compared with correlation processing methods of orthogonal signals.

**Originality**. For the first time the interference immunity in form of dependence of bit error probability on normalized signal/noise ratio of the data exchange by spread spectrum signals with variable entropy has been obtained.

**Practical value**. The obtained results prove the feasibility of using variable entropy spread spectrum signals data exchange method in the distributed telemetric information processing systems in specific circumstances.

**Keywords:** *spread spectrum signal, variable entropy, interference immunity, telemetry, mobile robot*